\pdfoutput=1
\documentclass[letterpaper]{article} 
\usepackage{aaai23}  
\usepackage{times}  
\usepackage{helvet}  
\usepackage{courier}  
\usepackage[hyphens]{url}  
\usepackage{graphicx} 
\usepackage{natbib}  
\usepackage{caption} 
\usepackage{algorithm}
\usepackage{algorithmic}

\usepackage{newfloat}
\usepackage{listings}
\DeclareCaptionStyle{ruled}{labelfont=normalfont,labelsep=colon,strut=off} 
\floatstyle{ruled}
\newfloat{listing}{tb}{lst}{}
\floatname{listing}{Listing}
\usepackage{amsfonts}
\usepackage{amsmath} 
\usepackage{amssymb}
\usepackage{bbding}
\usepackage{booktabs}
\usepackage{enumerate}
\usepackage{makecell}
\usepackage{microtype}
\usepackage{multirow}
\usepackage{nicefrac}
\usepackage{pifont}
\usepackage{xcolor}
\usepackage[switch]{lineno}

\title{Video Object of Interest Segmentation}
\author{
    Siyuan Zhou\textsuperscript{\rm 1}\thanks{Work done during an internship at Alibaba Group.}, Chunru Zhan\textsuperscript{\rm 2}, Biao Wang\textsuperscript{\rm 2}, Tiezheng Ge\textsuperscript{\rm 2}, Yuning Jiang\textsuperscript{\rm 2}, Li Niu\textsuperscript{\rm 1}\thanks{Corresponding author.}
}
\affiliations{
    \textsuperscript{\rm 1}MoE Key Lab of Artificial Intelligence, Shanghai Jiao Tong University, Shanghai, China\\
    \textsuperscript{\rm 2}Alibaba Group, Beijing, China\\

    ssluvble@sjtu.edu.cn, \{zhanchunru.zcr, eric.wb, tiezheng.gtz, mengzhu.jyn\}@alibaba-inc.com, coco235000@gmail.com
}

\usepackage{bibentry}

\begin{document}

\maketitle

\begin{abstract}

In this work, we present a new computer vision task named video object of interest segmentation (VOIS). Given a video and a target image of interest, our objective is to simultaneously segment and track all objects in the video that are relevant to the target image. This problem combines the traditional video object segmentation task with an additional image indicating the content that users are concerned with. Since no existing dataset is perfectly suitable for this new task, we specifically construct a large-scale dataset called LiveVideos, which contains 2418 pairs of target images and live videos with instance-level annotations. In addition, we propose a transformer-based method for this task. We revisit Swin Transformer and design a dual-path structure to fuse video and image features. Then, a transformer decoder is employed to generate object proposals for segmentation and tracking from the fused features. Extensive experiments on LiveVideos dataset show the superiority of our proposed method. 

\end{abstract}

\section{Introduction}\label{sec:introduction}

\begin{figure}[t]
\centering
\includegraphics[width=0.7\columnwidth]{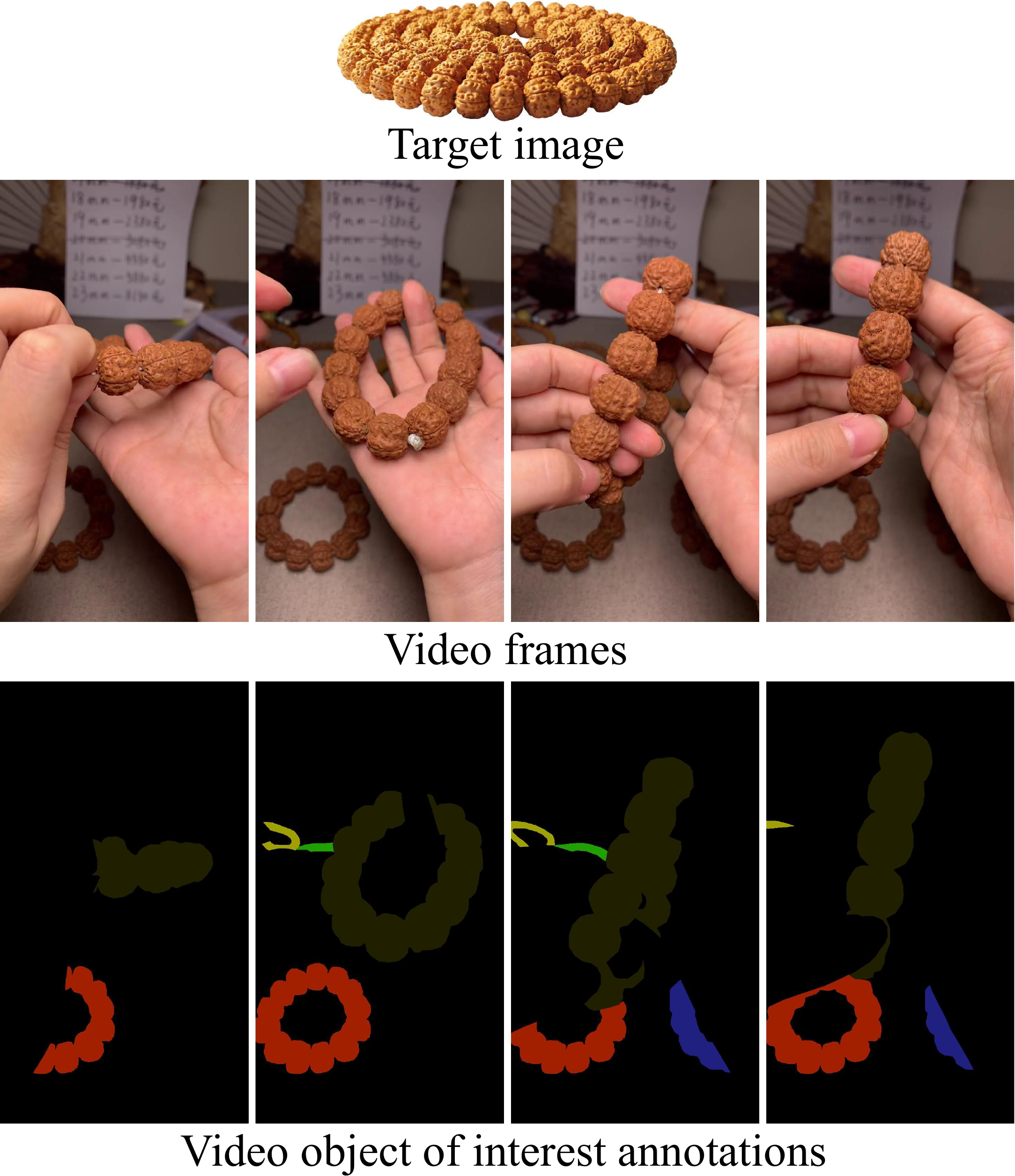}
\caption{Illustration of video object of interest segmentation (VOIS) with a pair of video clip and target image from LiveVideos dataset. The first row shows the target image. The next two rows display several video frames from the video clip, together with their corresponding VOIS annotations. Masks of the same color across frames belong to the same object.}
\label{fig:livevideos}
\end{figure}

Video object segmentation (VOS)~\cite{perazzi2016benchmark,pont20172017,xu2018youtube} refers to the task of segmenting class-agnostic object(s) in a video clip. It has been extensively studied and widely applied in various fields, like augmented reality, autonomous driving, video editing, \emph{etc}. Current researches on VOS has two main paradigms: unsupervised VOS~\cite{song2018pyramid,wang2019learning,zhou2020matnet,ren2021reciprocal} and semi-supervised VOS~\cite{caelles2017one,perazzi2017learning,voigtlaender2019feelvos,lin2019agss,bhat2020learning,liang2021video,mao2021joint,seong2021hierarchical,xie2021efficient}. The former one aims to automatically segment salient/primary objects while the latter one needs to segment objects specified by either human interaction or initial object annotation in the first frame. In this work, we propose a novel paradigm under the VOS task called Video Object of Interest Segmentation (VOIS). Different from the previous two settings, our new problem aims to simultaneously segment and track all objects in the video that are relevant to a given target image according to the user's interest, as well as requiring no additional annotation during inference. Each target image contains a single target object with white background. A video object is classified as a relevant object only if it looks like the same instance as the given target object in style, pattern, category, and color. Note that a relevant object with geometric deformation in the video is still considered as a relevant object \emph{w.r.t.} the given target object. Figure~\ref{fig:livevideos} illustrates a sample video with a target image and ground-truth annotations for the VOIS problem. The new paradigm could facilitate typical applications that require customized choices of objects for segmentation. For example, in advertising live broadcasts, the host may want to highlight the product that he/she is displaying. Under this condition, as long as the host offered a picture of the product in advance, the service provider could use VOIS techniques to obtain the segmentation of relevant products during the live broadcast, and then apply special effects to highlight these identified products in real time.

\begin{figure*}[t]
\centering
\includegraphics[width=1.8\columnwidth]{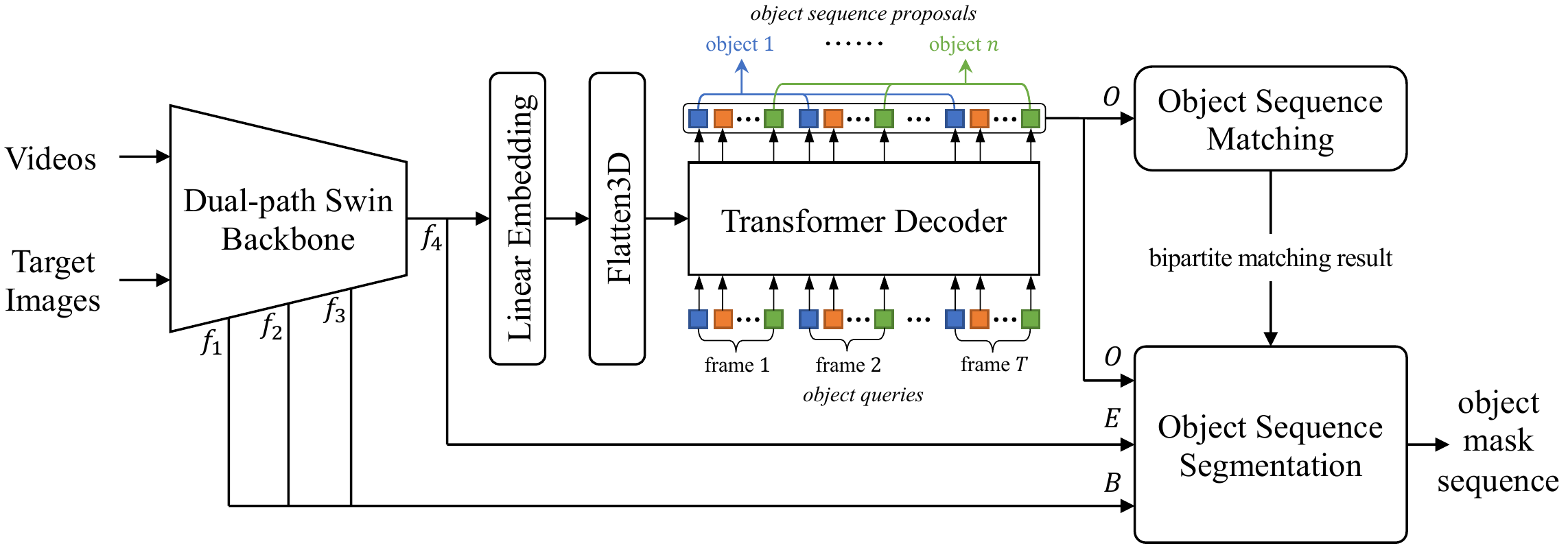}
\caption{The overall pipeline of our proposed method, which includes a dual-path Swin Transformer Backbone (see Section~\ref{subsec:method_step1}), a Transformer Decoder (see Section~\ref{subsec:method_step2}), and a object sequence matching/segmentation module (see Section~\ref{subsec:method_step3}).}
\label{fig:pipeline}
\end{figure*}

Video object of interest segmentation is intrinsically a multimodal problem that deals with both a video and an image input. Meanwhile, it requires simultaneous tracking and segmentation of multiple relevant objects in the video. The above two reasons make it more challenging than traditional VOS tasks. VOIS is also related to several existing segmentation tasks. For example, unsupervised video multi-object segmentation~\cite{ventura2019rvos,luiten2020unovost,zhou2020matnet,zhou2021target} aims to segment multiple class-agnostic salient objects in the video, but it can not deal with target objects specified by the user. Video instance segmentation~\cite{yang2019video,bertasius2020classifying,wang2021end,li2021spatial} aims to segment class-specific instances in the video, whereas VOIS can deal with objects of unseen categories during inference since the model learns class-agnostic knowledge in training. 

To our best knowledge, no previous work deals with video object of interest segmentation, and no existing video dataset is directly applicable to VOIS. Hence, we propose the first large-scale dataset for VOIS, called \emph{LiveVideos}. The new dataset contains 2003 high-resolution live videos and 2418 target images from the E-commerce live broadcast scenes, which constitute 2418 pairs of target image and video for training and inference. Meanwhile, our dataset includes annotations for 3341 video objects and 114k high-quality masks. Our new dataset could serve as a fundamental benchmark for not only video object of interest segmentation, but also traditional video object segmentation. The application scenarios of the dataset includes video retrieval, video highlight, \emph{etc}.

Furthermore, we propose a Transformer-based method for video object of interest segmentation. As illustrated in Figure~\ref{fig:pipeline}, the whole framework contains a dual-path Swin Transformer backbone, a Transformer decoder, and a object sequence matching and segmentation module. Swin Transformer~\cite{liu2021swin} has proved to be a well-performed general-purpose network for computer vision. We reuse and reconstruct Swin Transformer to be a dual-path backbone that accepts a 3D video input and a 2D image input simultaneously. Generally, the redesigned Swin Transformer functions as a backbone network that fuses the video feature and the target image feature, and outputs the attended video feature where video regions related to the target image are activated. After that, we introduce a Transformer decoder~\cite{vaswani2017attention,carion2020end} to extract object-level proposals from pixel-level backbone features. Finally, we adopt the instance sequence matching/segmentation module in VisTR~\cite{wang2021end} to arrange the object proposals according to ground-truth labels, and produce the segmentation masks for each object proposal. Extensive experiments on LiveVideos dataset demonstrate the effectiveness of our method.

In conclusion, the main contributions of this paper are:
\begin{itemize}
    \item We define and explore a new VOS paradigm called video object of interest segmentation (VOIS). 
    \item We create the first large-scale dataset for VOIS, containing 2418 pairs of target image and video clip from the live broadcast scenario.
    \item We propose an almost totally Transformer-based method for VOIS, and prove its advantages over several baselines. 
\end{itemize}

\begin{table*}[t]
\centering
\begin{tabular}{cccccc}
\toprule
Dataset & Video clips & Categories & Objects & Masks & Exhaustive \\
\midrule
FBMS~\cite{ochs2013segmentation} & 59 & 16 & 139 & 1.5k & \XSolidBrush \\
YouTubeObjects~\cite{jain2014supervoxel} & 96 & 10 & 96 & 1.7k & \XSolidBrush \\
DAVIS2016~\cite{perazzi2016benchmark} & 50 & - & 50 & 3.4k & \XSolidBrush \\
DAVIS2017~\cite{pont20172017} & 90 & - & 205 & 13.5k & \XSolidBrush \\
YouTubeVOS~\cite{xu2018youtube} & 4453 & 94 & 7755 & 197k & \XSolidBrush \\
YouTubeVIS~\cite{yang2019video} & 2883 & 40 & 4883 & 131k & \CheckmarkBold \\
LiveVideos & 2418 & - & 3341 & 114k & \CheckmarkBold \\
\bottomrule
\end{tabular}
\caption{Statistics of different video segmentation datasets.}
\label{table:datasets}
\end{table*}

\section{Related Work}\label{sec:relatedwork}

\paragraph{Video Object Tracking.}\label{para:vot}

Video object tracking methods either track objects based on the given bounding boxes in the first frame (\emph{i.e.}, detection-free tracking)~\cite{bertinetto2016fully,nam2016learning,feichtenhofer2017detect} or detect and track objects at the same time (\emph{i.e.}, detection-based tracking)~\cite{sadeghian2017tracking,wojke2017simple,son2017multi}. Both of them only require to produce bounding boxes, whereas video object of interest segmentation requires preciser segmentation masks. Besides, our task aims at objects specified by a target image, which also makes it different from video object tracking. Compared with video multi-object tracking~\cite{voigtlaender2019mots}, we have some additional differences: 1) our problem is not limited to moving objects, and 2) if an object goes out of scene for several frames then reappears, the object label should be consistent.

\paragraph{Video Object Segmentation.}\label{para:vos}

Video object segmentation (VOS) has two main settings: unsupervised VOS and semi-supervised VOS. The former one~\cite{ventura2019rvos,wang2019zero,lu2020learning,zhang2021deep} segments class-agnostic salient objects, while the latter one~\cite{oh2018fast,oh2019video,lu2020video,park2021learning,duke2021sstvos,ge2021video,hu2021learning} segments specified objects given in the first frame. Compared with the above two settings, our proposed video object of interest segmentation aims at segmenting video objects relevant to a specified target image. Meanwhile, video object of interest segmentation requires to simultaneously track different relevant objects, whereas traditional VOS mainly focuses on a single object.



\paragraph{Video Instance Segmentation.}\label{para:vis}

Video instance segmentation (VIS)~\cite{yang2019video,athar2020stem,wang2021end,wangtao2021end,lin2021video,liu2021sg} aims to simultaneously detect, track, and segment class-specific instances in a video clip. Compared with VIS, our proposed video object of interest segmentation has two main differences. First, our setting targets at class-agnostic objects, which means that it has the potential to segment unseen classes during inference. Second, the objects to segment is determined by a specified target image instead of a predefined class set, which makes it more flexible during application because users can arbitrarily determine what type of objects to segment by choosing the target image of interest.

\section{Video Object of Interest Segmentation}\label{sec:vois}

\paragraph{Problem Definition.}\label{para:def}

Given a target image and a video clip with $T$ frames, suppose there are $M$ video objects relevant to the target image. For the $i$-th object, we use $\mathbf{m}^{i}_{p\dots q}$ to denote its binary segmentation mask across the video, where $p$ and $q\in[1,T]$ represents its starting and ending time, respectively. Suppose a video object of interest segmentation algorithm produces $H$ object hypotheses (also called object sequence proposals). For the $j$-th hypotheses, the algorithm needs to produce a confidence score $s^j\in[0,1]$ and a sequence of predicted binary masks $\tilde{\mathbf{m}}^{j}_{\tilde{p}\dots\tilde{q}}$. The confidence score will be used in the evaluation metric.

Our objective is to minimize the difference between the hypotheses and the ground truth. It requires that a good VOIS algorithm should be able to 1) correctly detect relevant objects, 2) consistently track all relevant objects across frames, and 3) accurately segment all relevant objects.

\begin{figure*}[t]
\centering
\includegraphics[width=1.87\columnwidth]{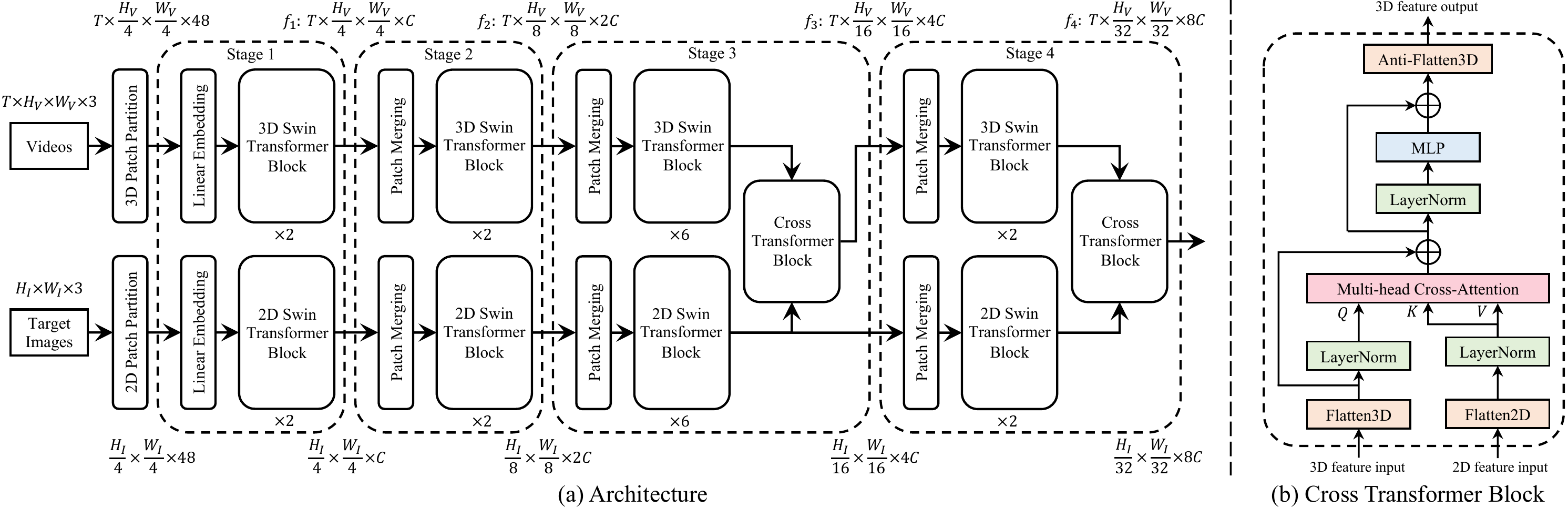}
\caption{(a) The architecture of our dual-path Swin Transformer backbone; (b) Structure of Cross Transformer block. MLP means multi-layer perceptron. Flatten3D means flattening spatial and temporal dimensions. Flatten2D means flattening spatial dimensions. Anti-Flatten3D means recovering the flattened spatial and temporal dimensions.}
\label{fig:backbone}
\end{figure*}

\paragraph{Evaluation Metrics.}\label{para:metric}

Since VIS~\cite{yang2019video} and VOIS share the same spirit of tracking and segmenting multiple objects simultaneously, we directly adapt the evaluation metrics (average precision AP and average recall AR) in VIS to our VOIS task. AP is defined as the area under the precision-recall curve, and AR is defined as the maximum recall given some fixed number of segmented objects per video. More detailed definitions are given in~\cite{yang2019video}. AP and AR work together to reflect the quality of hypotheses produced by the algorithm for evaluation. There is only one difference when we adapt the evaluation metrics to our problem. In VIS, AP and AR are calculated per category and then averaged across the category set. However, in our VOIS problem, since all objects are class-agnostic, AP and AR are directly calculated among all relevant video objects. In other words, evaluation metrics in VOIS can be regarded as a special case of those in VIS, where the category set has only one category, \emph{i.e.}, the relevant object.

\section{LiveVideos}\label{sec:livevideos}

Since no existing video segmentation dataset perfectly adapts to our video object of interest segmentation (VOIS) task, we need to establish a new benchmark for this task specifically. There are three important principles to satisfy when we establish the benchmark. The first concern is the source of data. As introduced in Section~\ref{sec:introduction}, advertising live broadcasts is a common and practical scenario where VOIS can be applied. This guides us to choose videos and target images from E-commerce live broadcast scenes to form our dataset. The second concern is the challenge of complex objects like occlusion, appearance change, frequent camera entry/leave, etc. We should take all these conditions into consideration to ensure the diversity and robustness of the dataset. The last concern is the quality of segmentation annotations. We should overcome the weakness of some existing datasets with polygon-based annotations.

Based on the above three principles, we establish a large-scale benchmark called \emph{LiveVideos}. We collect over 10k high-resolution live videos from the E-commerce live broadcast scenes and manually select 2003 representative videos from them. In the broadcast scenes of these selected videos, we collect 2418 relevant target images from the commodity banners, and ensure each of them is clear and not over-covered. Every target image contains a cropped target object with white-base background, indicating the commodity (\emph{e.g.}, clothing, jewellery, daily supplies) displayed in one of the 2003 live videos. Therefore, we have 2418 pairs of live videos and target images. For each pair, we carefully clip out one video clip with a duration of 5.0$\sim$7.2 seconds from the live video, and manually verify that this clip contains the correct target object(s) and is useful for our task (\emph{e.g.}, not too blurry or shaky, no scene transition). After the 2418 video clips are selected, we ask professional human annotators to annotate all the objects (no more than 10 in fact) in each video clip that are relevant to the corresponding target image. We follow~\cite{xu2018youtube} to adopt a skip-frame annotation strategy. The annotation is performed every four frames in a 20fps frame rate, resulting in a 5fps sampling rate, so no more than 36 frames are annotated in each video clip. Some annotation examples are shown in Figure~\ref{fig:livevideos}. As a result, our LiveVideos dataset contains 2418 pairs of video clips and target images, and 3341 video objects with 114k high-quality object masks, which form a large-scale benchmark. Table~\ref{table:datasets} compares LiveVideos with some existing video segmentation datasets. It shows that the scale of our dataset is comparable with YouTubeVOS~\cite{xu2018youtube} and YouTubeVIS~\cite{yang2019video}, and evidently larger than other commonly used datasets.

\section{Methodology}\label{sec:methodology}

Video object of interest segmentation (VOIS) task takes a video clip and a target image as input, and aims to track and segment all video objects that are relevant to the target image. Generally, we tackle the VOIS task with three steps, as shown in Figure~\ref{fig:pipeline}. First, we design a dual-path Swin Transformer to fuse video features and image features in Section~\ref{subsec:method_step1}. Second, we employ a Transformer decoder to generate object proposals from the fused features in Section~\ref{subsec:method_step2}. Third, we use a sequence matching module to arrange the object proposals and a sequence segmentation module to produce the segmentation results for each object proposal in Section~\ref{subsec:method_step3}.

\subsection{Dual-path Swin Transformer}\label{subsec:method_step1}

2D Swin Transformer~\cite{liu2021swin} was proposed as a general-purpose backbone to extract image features with a totally end-to-end Transformer-based network. Specifically, it splits the input image into non-overlapping 2D patch tokens and applies four stages to process these tokens. Each stage contains a predefined number of consecutive Swin Transformer Blocks with the proposed 2D window based multi-head self-attention modules (W-MSA) or 2D shifted-window based multi-head self-attention modules (SW-MSA). Four stages work together to produce a hierarchical representation as output. 3D Swin Transformer~\cite{liu2022video} extends the 2D version to deal with video inputs. Likewise, it splits videos into 3D patch tokens, and changes the 2D window based attention modules into 3D versions.

In this work, we combine the 2D version and the 3D version to form a dual-path Swin Transformer that accepts both an image and a video input, as shown in Figure~\ref{fig:backbone}. The target image and the video are defined with size $H_{\mathrm{I}}\times W_{\mathrm{I}}\times 3$ and $T\times H_{\mathrm{V}}\times W_{\mathrm{V}}\times 3$, respectively. We use different subscripts $\mathrm{I}$/$\mathrm{V}$ in image/video to avoid confusion. The video has an extra dimension $T$ indicating it contains $T$ frames. We treat each 2D patch of size $4\times 4\times 3$ as token in the 2D path, and each 3D patch of size $1\times 4\times 4\times 3$ as token in the 3D path. The 2D patch partitioning layer obtains $\frac{H_{\mathrm{I}}}{4}\times\frac{W_{\mathrm{I}}}{4}$ 2D patches and the 3D patch partitioning layer obtains $T\times\frac{H_{\mathrm{V}}}{4}\times\frac{W_{\mathrm{V}}}{4}$ 3D patches. Each patch/token consists of a 48-dimensional feature. Then a 2D/3D linear embedding layer is employed to map the 2D/3D token to an arbitrary dimension $C$.

Like the traditional 2D/3D Swin Transformer, The dual-path Swin Transformer architecture also contains four stages, with each stage combining a 2D stage and 3D stage. Under this design, each stage accepts dual-path inputs and generates dual-path outputs. In the 2D/3D patch merging layer of each stage, the height dimension and the width dimension are down-sampled, while the token dimension doubles. Note that we follow the prior work~\cite{liu2022video} not to down-sample along the temporal dimension in the 3D path. The designs of each 2D/3D Swin Transformer block remain the same as~\cite{liu2021swin,liu2022video}, so we omit the details here.

Since we aim to find video objects of interest conditioned on the target image, we add a Cross Transformer block to fuse video and image feature in stage 3 and stage 4, respectively. The Cross Transformer block contains a multi-head cross-attention and a MLP. We first apply the multi-head cross-attention by treating video feature as query and image feature as key/value. Then, we handle the attended feature via a 2-layer MLP. The output feature of the Cross Transformer block functions as the 3D video input of the next stage. Under this design, video regions relevant to the target image tend to be activated, while the remaining parts tend to be deactivated. Then, the following steps would pay more attention to the video regions that we are interested in, which facilitates segmentation of relevant objects in an implicit way.

For ease of representation, we use $f_1$, $f_2$, $f_3$, and $f_4$ to represent the output 3D video features of each stage, as shown in both Figure~\ref{fig:pipeline} and Figure~\ref{fig:backbone}. We finally obtain a 3D video feature $f_4$ of size $T\times \frac{H_{\mathrm{V}}}{32}\times \frac{W_{\mathrm{V}}}{32}\times 8C$ as the backbone output.

\begin{table*}[t]
\centering
\begin{tabular}{ccccccc}
\toprule
Method & Backbone & AP & AP$_{50}$ & AP$_{75}$ & AR$_1$ & AR$_{10}$ \\
\midrule
MaskTrack R-CNN~\cite{yang2019video} & ResNet-50~\cite{he2016deep} & 29.0 & 46.4 & 32.1 & 32.5 & 35.5 \\
VisTR~\cite{wang2021end} & ResNet-50~\cite{he2016deep} & 34.9 & 54.1 & 37.0 & 38.3 & 41.8 \\
VisTR~\cite{wang2021end} & ResNet-101~\cite{he2016deep} & 37.0 & 56.4 & 39.2 & 38.1 & 43.7 \\
\midrule
Ours & Dual-path Swin Transformer & 38.8 & 61.6 & 41.8 & 41.2 & 46.6 \\
\bottomrule
\end{tabular}
\caption{Quantitative comparison between different methods.}
\label{table:sota}
\end{table*}

\begin{figure*}[t]
\centering
\includegraphics[width=1.67\columnwidth]{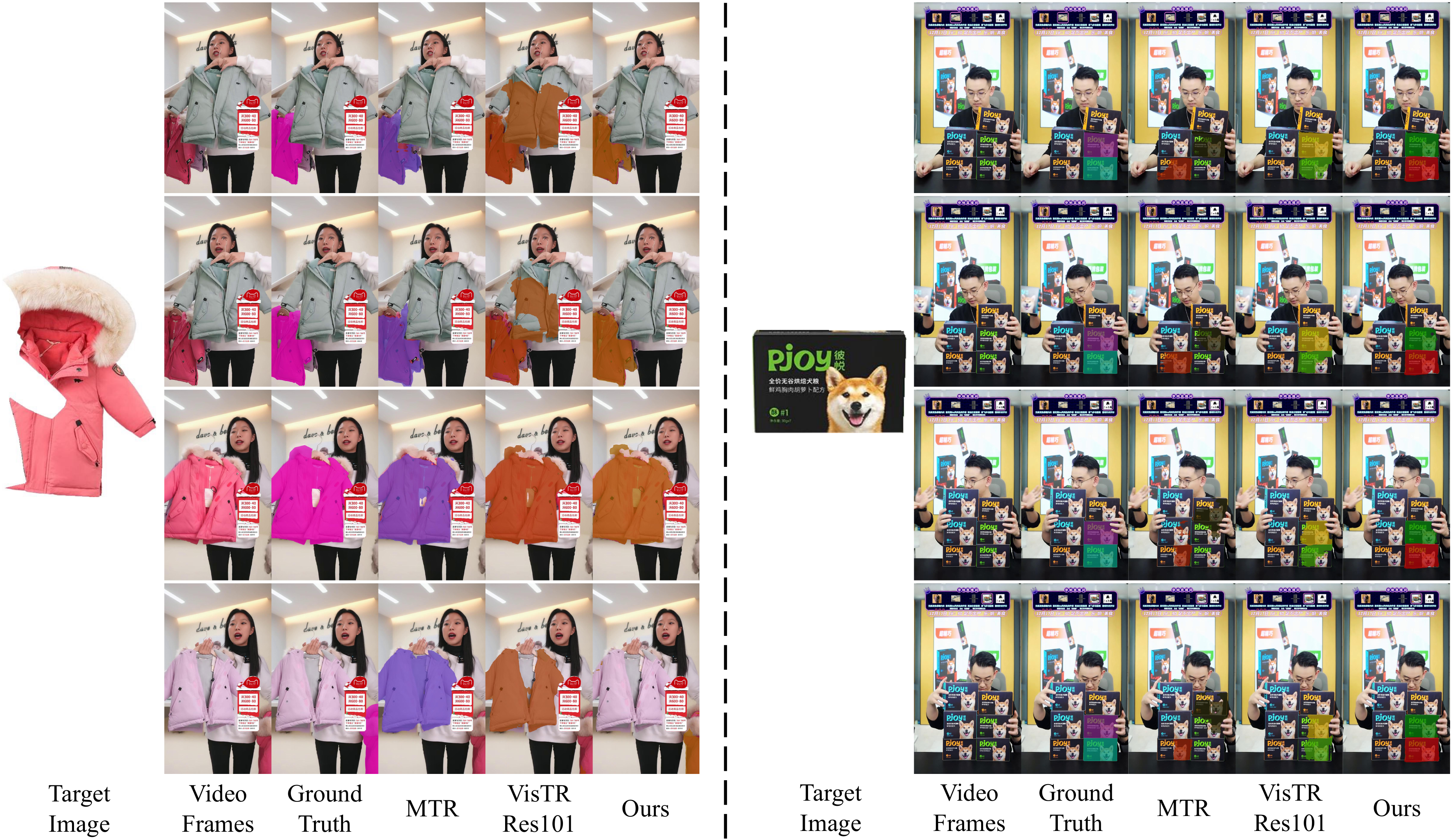}
\caption{Visualization of video object of interest predictions of different methods. MTR is short for MaskTrack R-CNN. For each method, segmentation masks of the same color across different frames belong to the same object. Zoom in for more details.}
\label{fig:vois_result}
\end{figure*}

\subsection{Transformer Decoder}\label{subsec:method_step2}

Motivated by DETR~\cite{carion2020end}, we incorporate a Transformer decoder to decode pixel-level features into object-level representations for each frame. As illustrated in Figure~\ref{fig:pipeline}, before the backbone feature $f_4$ enters the Transformer decoder, we apply a linear embedding layer on $f_4$ to map it from backbone dimension to decoder hidden dimension. Then, we flatten its spatial and temporal dimensions so that it could be fed into the Transformer decoder. During implementation, we introduce a fixed number of input embeddings to query object-level features, termed as object queries. Specifically, the model decodes $n$ objects for each frame, so the total number of object queries is $N=n\cdot T$. The Transformer decoder works by taking the output of the dual-path Swin Tranformer backbone and $N$ object queries as input to produce $N$ object-level features. After that, $N$ object-level features form $n$ object sequence proposals (\emph{abbr.}, object proposals), and each object proposal is composed by $T$ object-level features from the same index of different frames. The workflow of Transformer Decoder is shown in Figure~\ref{fig:pipeline}. We denote the decoder output by $O$, which represents the set of $n$ object sequence proposals.

\subsection{Object Sequence Matching and Segmentation}\label{subsec:method_step3}

Once we obtain object sequence proposals, we will match them with the ground truth object sequences via a object sequence matching module. Then, we predict the mask sequence for each object proposal. After that, we calculate losses between the predicted mask sequence and its matched ground truth sequence to optimize the model, which is achieved by a object sequence segmentation module.

We draw inspirations from VisTR~\cite{wang2021end} to realize object sequence matching and segmentation. VisTR deals with the video instance segmentation (VIS) task. It first matches its predicted instance sequence proposals with the ground truth sequences via an instance sequence matching module with \emph{bipartite matching loss}. Then, it segments each instance sequence proposal and optimizes the model via an instance sequence segmentation module with a \emph{Hungarian loss}. The above steps of VisTR is somewhat similar to our task, which guides us to adapt the instance sequence matching/segmentation module in VisTR to our required object sequence matching/segmentation module. Generally, there exists two differences between object sequence matching/segmentation and instance sequence matching/segmentation.

The first difference lies in the formation of features. As shown in Figure~\ref{fig:pipeline}, our object sequence segmentation module follows the instance sequence segmentation module in VisTR to accept three feature inputs: decoder feature $O$, encoder feature $E$, and backbone feature $B$. In our implementation, $O$ remains output feature of the Transformer decoder. $E=f_4$ is exactly the output feature of stage 4 in our dual-path Swin Transformer backbone. $B=\{f_1,f_2,f_3\}$ is the set of multi-level features from the beginning three stages in our dual-path Swin Transformer backbone. The second difference lies in class labels. VIS is a multi-class problem and has a predefined category set. Differently, our VOIS task aims at class-agnostic objects, \emph{i.e.}, single-class objects. Therefore, we modify the category-relevant loss terms in \emph{bipartite matching loss} and \emph{Hungarian loss} from multi-class forms to their single-class counterparts. The \emph{Hungarian loss} contains three parts: classification, box regression, and segmentation. The classification / segmentation part produces the confidence scores / binary masks for object proposals respectively, which are both necessary outputs required by our VOIS task. The detailed loss forms remain unchanged, so we omit here.

\section{Experiment}\label{sec:experiment}

\subsection{Dataset and Evaluation Metrics}\label{subsec:dataset_metric}

We conduct experiments on LiveVideos dataset. As introduced in Section~\ref{sec:livevideos}, we have 2418 pairs of live videos and target images (\emph{i.e.} 2418 samples) to form the whole LiveVideos dataset. We then randomly split the dataset into 1935 training samples and 483 test samples. Each sample is annotated with pixel-level segmentation masks and object labels of all the objects that are relevant to the corresponding target image. We train models on the training set and evaluate them on the test set. The evaluation metrics are Average Precision (AP) and Average Recall (AR) as introduced in Section~\ref{para:metric}.

\subsection{Implementation Details}\label{subsec:details}

\paragraph{Network Structure.}\label{para:network}

As introduced in Section~\ref{sec:livevideos}, each video clip contains no more than 36 frames, so we set the input video sequence length $T$ as 36. The dual-path Swin Transformer backbone is a fusion of 2D Swin Transformer~\cite{liu2021swin}, 3D Swin Transformer~\cite{liu2022video} with temporal patch size modified to 1, and two Cross Transformer blocks. The tiny version of 2D/3D Swin Transformer is chosen due to GPU memory limitation. The initial token dimension $C$ is 96, so the backbone output dimension is $8C=768$. Each Cross Transformer block contains a multi-head attention~\cite{vaswani2017attention} and a MLP, with short-cut connections. MLP is composed of two linear layers, with GELU non-linearity in between. The Transformer decoder follows the structure in DETR~\cite{carion2020end}, which contains 6 decoder layers with the hidden dimension modified to 384. The Transformer decoder decodes $n=10$ objects for each frame. The linear embedding layer between the Swin backbone and the Transformer decoder is a linear layer that maps the backbone dimension $768$ to the decoder hidden dimension $384$. The object sequence matching/segmentation module follows the design in VisTR~\cite{wang2021end}, with the classification head modified from 41 classes (40 YouTube-VIS categories and background category) to 2 classes (relevant category and background category). The other hyper-parameters in the backbone network (\emph{resp.}, Transformer Decoder, object sequence matching/segmentation) follow the default settings in Swin Transformer (\emph{resp.}, DETR, VisTR).

\paragraph{Data Preprocessing.}\label{para:preprocess}

We first augment the input videos and target images with random horizontal flip and random crop. Then, we resize their shorter edges to 224 by keeping the aspect ratio unchanged. Finally, we apply normalization before feeding them into the network. 

\paragraph{Optimization.}\label{para:optimization}

We adopt AdamW~\cite{loshchilov2017decoupled} optimizer with learning rate being $10^{-5}$ for the dual-path Swin Transformer backbone and $10^{-4}$ for the remaining parts. The model is trained with 18 epochs, where the learning rate decays by 10x after 12 epochs. We initialize the backbone network with the weights of Swin Transformer pretrained on ImageNet~\cite{deng2009imagenet}, and initialize the Transformer decoder with weights of DETR pretrained on MS COCO~\cite{lin2014microsoft}. The model is trained on 32 Tesla V100 GPUs with distributed parallel. Each GPU card deals with one pair of video clip and target image in one batch. We perform inference on a single V100 GPU, and retain object proposals with confidence scores larger than 0.001. Experiments are conducted with PyTorch-1.7~\cite{paszke2019pytorch}.

\subsection{Baselines.}\label{subsec:baseline}

To our best knowledge, no existing method directly adapts to our task. Therefore, we absorb ideas from related tasks to form baselines. Video instance segmentation (VIS)~\cite{yang2019video} is similar to our task in that it also requires to track and segment multiple objects. To adapt a VIS method to a VOIS method, the modification consists of three aspects. First, we should modify the input of the network so that it could accept two inputs, \emph{i.e.}, a video input and an image input. Second, we should fuse video features and image features at a typical stage in the network. Third, we should change the multi-class classification head of the network to its single-class counterpart to deal with class-agnostic output. In this work, we choose to adapt from two representative VIS methods to form our baselines: MaskTrack R-CNN~\cite{yang2019video} and VisTR~\cite{wang2021end}.

\paragraph{MaskTrack R-CNN.}\label{para:mtr}

MaskTrack R-CNN~\cite{yang2019video} absorbs the `tracking-by-detection' idea from multi-object tracking to form its method. Typically, it add a tracking head with an external memory into the classical Mask R-CNN to track object instances across frames. To adapt MaskTrack R-CNN~\cite{he2017mask} to our task, we add a secondary ResNet~\cite{he2016deep} backbone to accept target image input. Then, we use a Cross Transformer block to fuse video frame features and target image features outputted by two backbones. The fused features are sent to the original network structures after the original video backbone.

\begin{table}[t]
\centering
\begin{tabular}{cp{6mm}<{\centering}p{6.5mm}<{\centering}p{6.5mm}<{\centering}p{6mm}<{\centering}p{6.5mm}<{\centering}}
\toprule
Target image path & AP & AP$_{50}$ & AP$_{75}$ & AR$_1$ & AR$_{10}$ \\
\midrule
& 26.7 & 42.9 & 28.1 & 31.0 & 38.0 \\
\Checkmark & 38.8 & 61.6 & 41.8 & 41.2 & 46.6 \\
\bottomrule
\end{tabular}
\caption{Ablation study on utility of target image path.}
\label{table:ab_path}
\end{table}

\begin{table}[t]
\centering
\begin{tabular}{ccp{6mm}<{\centering}p{6.5mm}<{\centering}p{6.5mm}<{\centering}p{6mm}<{\centering}p{6.5mm}<{\centering}}
\toprule
Stage 3 & Stage 4 & AP & AP$_{50}$ & AP$_{75}$ & AR$_1$ & AR$_{10}$ \\
\midrule
\Checkmark & & 37.5 & 60.3 & 41.0 & 39.9 & 44.6 \\
& \Checkmark & 37.7 & 59.8 & 41.4 & 39.5 & 45.5 \\
\Checkmark & \Checkmark & 38.8 & 61.6 & 41.8 & 41.2 & 46.6 \\
\bottomrule
\end{tabular}
\caption{Ablation study on position of Cross Transformer.}
\label{table:ab_fuse}
\end{table}

\paragraph{VisTR.}\label{para:vistr}

VisTR~\cite{wang2021end} is the first Transformer-based VIS method that treats VIS as a direct end-to-end parallel sequence prediction problem. To adapt VisTR to our task, we similarly add a secondary ResNet~\cite{he2016deep} backbone to accept target image input. Then we add a Cross Transformer block after each Transformer encoder layer to fuse features. The Cross Transformer block takes the output video feature of the current encoder layer and the target image feature from secondary backbone as input to produce a fused feature, which functions as the input video feature of the next encoder layer. Since there six Transformer encoder layers, we have six Cross Transformer blocks correspondingly. The output of the final Cross Transformer block is sent to the instance sequence matching/segmentation module.

\subsection{Main Results.}\label{subsec:result}

Table~\ref{table:sota} presents the comparison results between baseline methods and our proposed method. Generally speaking, our proposed method achieves the best results among different methods. In detail, our method surpasses MaskTrack R-CNN by 9.8 AP / 8.7 AR$_1$, and surpasses VisTR with ResNet-101 backbone by 1.8 AP / 3.1 AR$_1$. It proves that our method achieves not only better mask quality, but also better temporal consistency and relevant object detection rate. It is noteworthy that although our chosen backbone is merely the tiny version of Swin Transformer, our proposed method still outperforms baselines with ResNet backbones, which implies the great potentiality of our method.

Figure~\ref{fig:vois_result} displays two example cases predicted by different methods. As illustrated, our proposed method performs better in the several challenging scenarios: 1) relevant object(s) with heavy motion, 2) relevant object(s) surrounded by multiple confusing objects that are easily misidentified. For the first scenario (left side of Figure~\ref{fig:vois_result}), our method accurately tracks and segments the relevant object (\emph{i.e.} the pink overcoat) despite its long-distance movement. For the second scenario (right side of Figure~\ref{fig:vois_result}), our method precisely locates and segments relevant objects with the correct color (\emph{i.e.} boxes with green patterns). Meanwhile, our method seamlessly predicts finer segmentation boundaries. These quantitative results further verify the effectiveness of our proposed method.

\subsection{Ablation Studies.}\label{subsec:ablation}

\paragraph{Utility of Target Image Path.}\label{para:imagepath}

To prove that our architecture effectively studies the target image information and use it to identify relevant objects in the video clip, we perform an experiment without the target image path. Specifically, the 2D Swin Transformer path and the two Cross Transformer blocks in Figure~\ref{fig:backbone} are deleted. We simply use the 3D Swin Transformer path to extract video frame features, which are then sent to the Transformer decoder. The comparison results are shown in Table~\ref{table:ab_path}. Unsurprisingly, the network witnesses a sharp performance drop without the target image input. In detail, AP and AR$_1$ decrease by 12.1 and 10.2, respectively. The reason for this phenomenon is intuitive. When we do not incorporate target images, the network does not know what object(s) to identify and segment. As a result, it would tend to randomly identify objects in videos during inference, which severally harms its ability to detect user-specified objects.

\paragraph{Position of Cross Transformer Block.}\label{para:fuseposition}

In our default implementation, we adopt two Cross Transformer blocks in stage 3 and stage 4, as shown in Figure~\ref{fig:backbone}. There are two reasons why we do not place Cross Transformer blocks in the beginning two stages. First, features in the beginning two stages are low-level features in shallow network layers, so they are not expressive enough for the model to find relevant information from the target image. Second, features in the beginning two stages have comparably large spatial dimensions, which makes it space-consuming to compute attention weights in the Cross Transformer block. Considering the above two points, we only use Cross Transformer in the deeper two layers. Table~\ref{table:ab_fuse} presents a comparison to examine whether each Cross Transformer block in stage 3 / stage 4 is necessary for the overall performance gain. Deleting the Cross Transformer block in stage 3 (\emph{resp.}, stage 4) gives rise to the performance drop of 1.1 AP / 0.7 AR$_1$ (\emph{resp.}, 1.3 AP / 0.3 AR$_1$). The comparison shows that both Cross Transformer blocks contribute to the final segmentation results, so we maintain both of them as our default setting.

\section{Conclusion}\label{sec:conclusion}

In this work, we present a new task named video object of interest segmentation (VOIS) and specifically construct a large-scale dataset called Livevideos for this task. We also propose an end-to-end Transformer-based method to deal with this multi-modal problem. The proposed method is proved to perform well and surpass several baselines. Compared with the traditional VOS tasks, our proposed VOIS task adopts an additional easily available target image as input to specify what kind of object(s) to track and segment in a video, which makes it user-friendly and conveniently applicable in many situations. We believe this work could attract and promote future research on video object of interest segmentation.

\section*{Acknowledgments}\label{sec:acknowledgments}

The work is supported by the National Science Foundation of China (62076162), the Shanghai Municipal Science and Technology Major/Key Project, China (2021SHZDZX0102, 20511100300), and Alibaba Group through Alibaba Innovation Research Program.

\bibliography{ref}

\end{document}


\maketitle

\noindent In this supplementary file, we first compare the model efficiency of different methods in Section~\ref{sec:efficiency}. Then, we perform a significance test in Section~\ref{sec:significance}. After that, we show more visualizations in Section~\ref{sec:visualize}. We also analyze the limitations of our method in Section~\ref{sec:limit}. Finally, we introduce more details of our proposed LiveVideos dataset in Appendix~\ref{appsec:dataset}.

\section{Model Efficiency}\label{sec:efficiency}

In Section 6.4 in the main paper, we have compared the performance between baselines and our method, and have shown the superiority of our method. In this section, we aim to further compare the model efficiency between the strongest baseline VisTR~\cite{wang2021end} and our method. We adopt two metrics for comparison. The first metric is the number of proceeded frames per second (FPS) of the model without considering the data loading process. The second metric is the number of learnable parameters of the model. For fairness, we use the same spatial resolution $224\times 224$ for video/image input in different methods. The comparison results are listed in Table~\ref{table:efficiency_supp}, which show that our proposed method achieves a higher FPS (93.3 \emph{v.s.} 87.1/74.7) during inference even when our model has the fewest parameters (89.7M \emph{v.s.} 97.5M/135.3M). Without bells and whistles, our method achieves the highest model efficiency.

\begin{table*}[t]
\centering
\begin{tabular}{ccccccc}
\toprule
Method & Backbone & Input spatial resolution & FPS w/o data loading & Params (M) \\
\midrule
VisTR~\cite{wang2021end} & ResNet-50~\cite{he2016deep} & $224\times 224$ & 87.1 & 97.5 \\
VisTR~\cite{wang2021end} & ResNet-101~\cite{he2016deep} & $224\times 224$ & 74.7 & 135.3 \\
\midrule
Ours & Dual-path Swin Transformer & $224\times 224$ & 93.3 & 89.7 \\
\bottomrule
\end{tabular}
\caption{Efficiency comparison between different methods. We adopt two metrics: 1) the number of proceeded frames per second (FPS) without considering data loading, and 2) the number of learnable model parameters. The experiments are implemented by Pytorch 1.7 with input spatial resolution $224\times 224$ and batch size 1 on a single 16G Tesla P100 GPU.}
\label{table:efficiency_supp}
\end{table*}

\section{Significance Test}\label{sec:significance}
    
For our main experiments in Table 2 in the main paper, we set the default seed as 42 in all experiments to determine the randomly initialized parameters in models. In this section, we reset the seed to 10 different values, and run the strongest baseline VisTR~\cite{wang2021end} with ResNet-101~\cite{he2016deep} backbone and our method for 10 times with these different seeds, respectively. As a result, AP is $\rm Mean\pm Std=37.0\pm 0.2$ in VisTR and $\rm 38.9\pm 0.1$ in our method. Meanwhile, AR$_1$ is $\rm Mean\pm Std=38.0\pm 0.1$ in VisTR and $\rm 41.2\pm 0.1$ in our method. At the significant level 0.05, we perform significance test on both methods. The p-values w.r.t. AP / AR$_1$ are 9.9e-3 / 1.4e-2 for VisTR and 8.1e-3 / 9.2e-3 for our method, which are all below 0.05. This proves that the superiority of our method is statistically significant. 

\section{Visualizations}\label{sec:visualize}

\begin{figure*}[t]
\centering
\includegraphics[width=1.95\columnwidth]{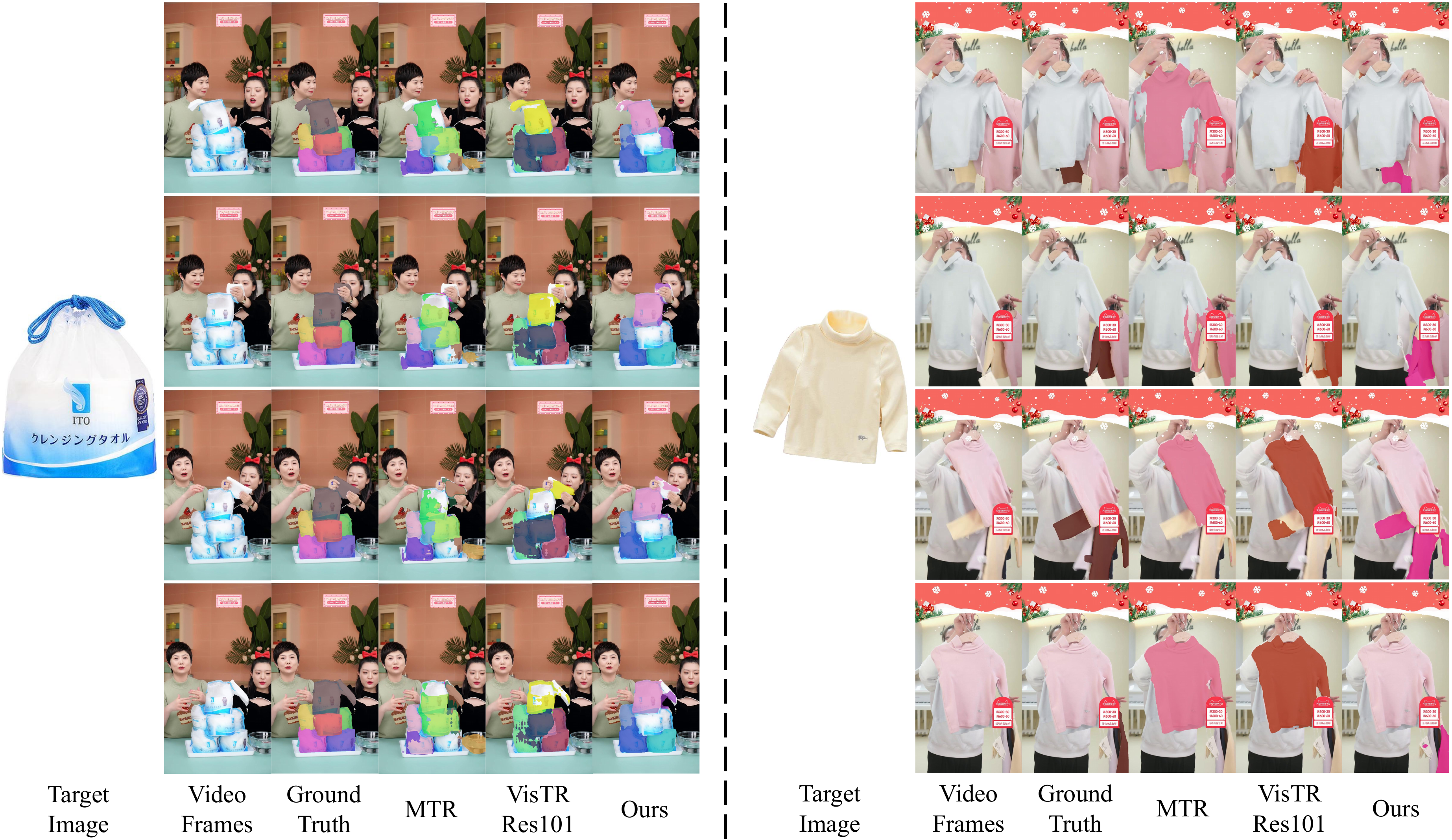}
\caption{Visualization (part-1) of VOIS predictions of different methods. MTR is short for MaskTrack R-CNN. For each method, segmentation masks of the same color across different frames belong to the same object. Zoom in for more details.}
\label{fig:vois_result_supp1}
\end{figure*}

\begin{figure*}[t]
\centering
\includegraphics[width=2.03\columnwidth]{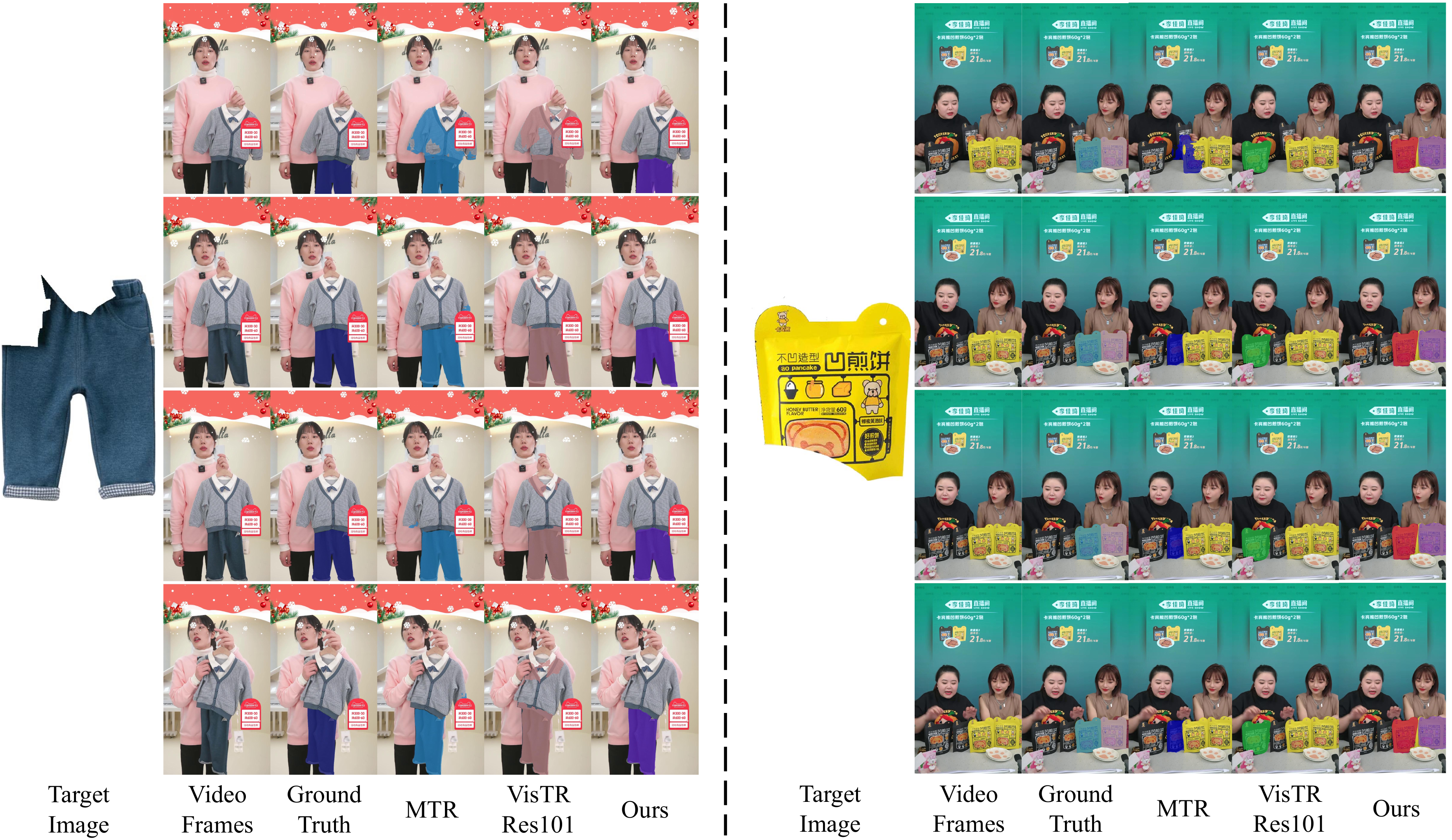}
\caption{Visualization (part-2) of VOIS predictions of different methods. MTR is short for MaskTrack R-CNN. For each method, segmentation masks of the same color across different frames belong to the same object. Zoom in for more details.}
\label{fig:vois_result_supp2}
\end{figure*}

In Figure~\ref{fig:vois_result_supp1} and Figure~\ref{fig:vois_result_supp2}, we provide more visualizations of MaskTrack R-CNN~\cite{yang2019video}, VisTR~\cite{wang2021end}, and our method, which supplement Figure 4 in the main paper. These visualizations show four advantages of our method against baselines as follows:
\begin{enumerate}[(a)]
    \item In the left case of Figure~\ref{fig:vois_result_supp1}, our method works better in identifying different relevant objects and tracing their boundaries when multiple similar objects gather together.
    \item In the right case of Figure~\ref{fig:vois_result_supp1}, our method can differentiate the heavily obscured relevant object from the confusing foreground objects.
    \item In the left case of Figure~\ref{fig:vois_result_supp2}, our method intelligently and precisely locates the region of the relevant object without hesitating about the adjacent related regions. 
    \item In the right case of Figure~\ref{fig:vois_result_supp2}, our method performs better in identifying the correct object color specified by the target image.
\end{enumerate}
The above advantages further prove the effectiveness of our proposed method in video object of interest segmentation.

\section{Limitations}\label{sec:limit}

\begin{figure*}[t]
\centering
\includegraphics[width=2.11\columnwidth]{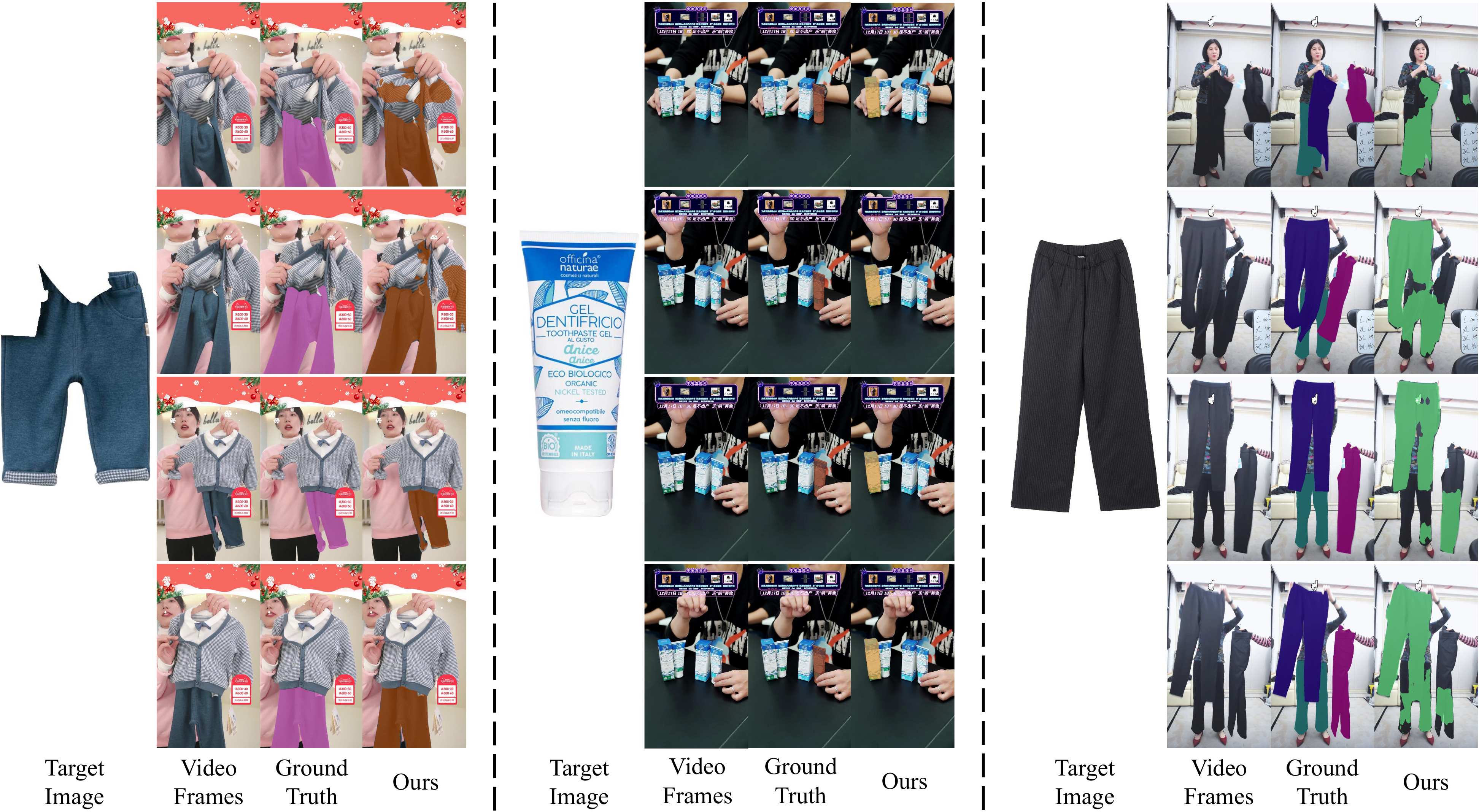}
\caption{Visualization of failure cases of our method. In each case, segmentation masks of the same color across different frames belong to the same object. Zoom in for more details.}
\label{fig:failure_case_supp}
\end{figure*}

In Figure~\ref{fig:failure_case_supp}, we show three failure cases of our method. Based on each case, we analyze the potential problems of our method and provide possible solutions correspondingly:
\begin{enumerate}[(a)]
    \item In the left case of Figure~\ref{fig:failure_case_supp}, our method fails in the first multiple frames whereas succeeds in the latter frames. We suppose that if the model knows which frame(s) it performs well from an external module, it could then use the predictions of these well-performed frames to purify those of badly-performed ones. Therefore, our improvement direction is to design and learn such a module, and use it to post-process video frames during inference. 
    \item In the middle case of Figure~\ref{fig:failure_case_supp}, our method fails to locate the correct relevant objects when multiple objects have similar colors/patterns in their body parts. To solve this problem, the model should be designed to pay more attention to the boundary shapes of different object proposals so as to get rid of confused ones.
    \item In the right case of Figure~\ref{fig:failure_case_supp}, our method fails to simultaneously detect multiple relevant objects when their appearances are similar and boundaries are obscure. This disadvantage could be alleviated via inspiration from existing image/video instance segmentation methods since they have proven to be proficient in splitting apart different instances of the same class.
\end{enumerate}
The above three perspectives shed light on improvement directions for future researches on video object of interest segmentation. We will also work on these directions later.

\appendix

\section{LiveVideos}\label{appsec:dataset}

\begin{figure*}[t]
\centering
\includegraphics[width=1.8\columnwidth]{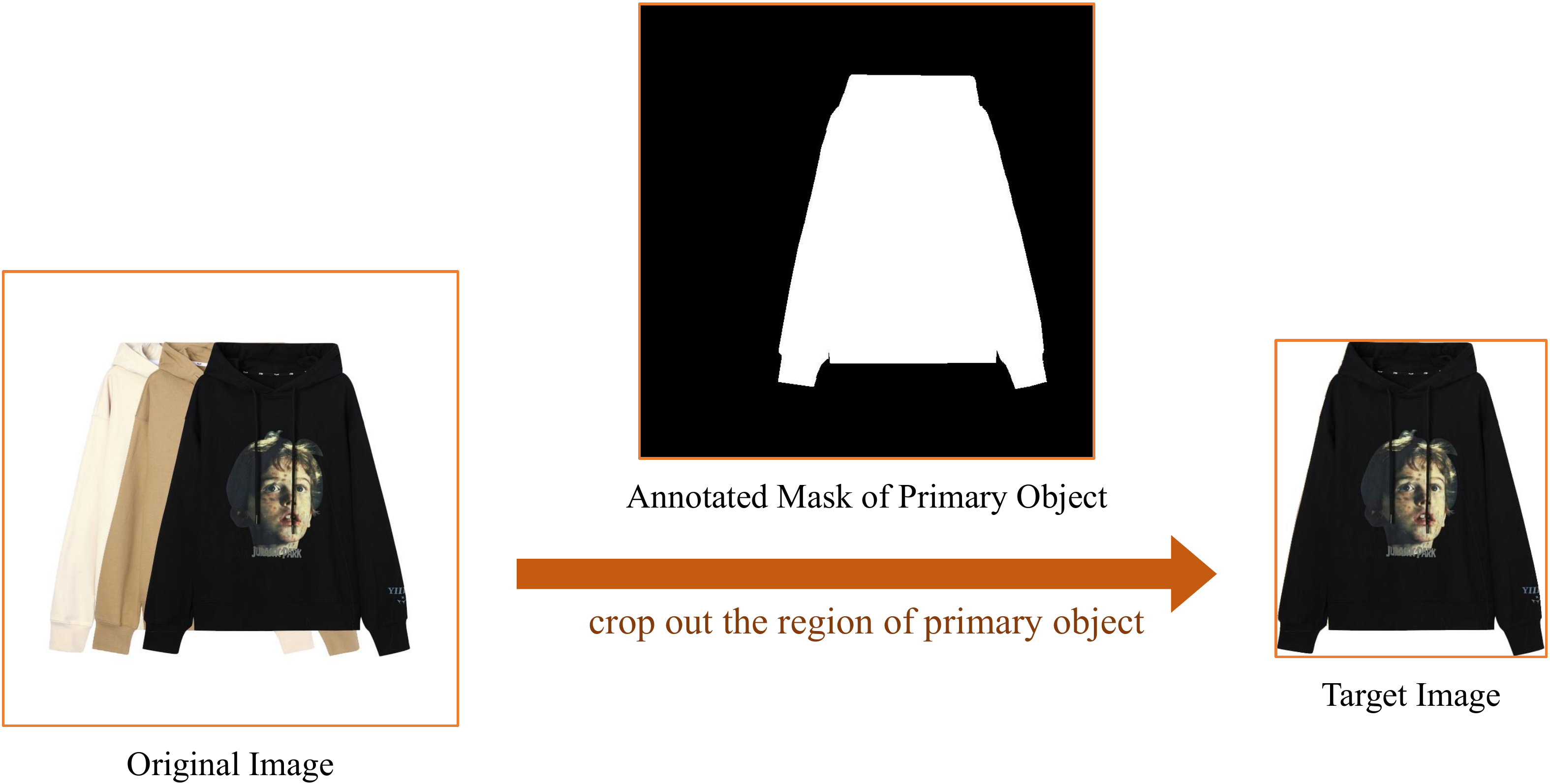}
\caption{Illustration of obtaining a target image from the original image.}
\label{fig:target_supp}
\end{figure*}

In Section 4 in the main paper, we have comprehensively introduced our LiveVideos dataset. In this data appendix, we will discuss about the process of obtaining the target images. As we have mentioned in Section 4 in the main paper, we collect relevant target images from the commodity banners in the broadcast scenes of our selected live videos. In fact, these collected images usually contains multiple objects and have white margins, as illustrated in the left side of Figure~\ref{fig:target_supp}. To make the learning process simple, we would like each target image only contains a single object in our dataset. Therefore, we ask professional annotators to annotate the segmentation mask of the primary object in each collected image, and manually ensure that the annotated object indeed exists in the video clip corresponding to this image. After that, we use the annotated mask to crop out the target object from the original image, and finally obtain the target image we use in the dataset. The above process is illustrated in Figure~\ref{fig:target_supp}.

\bibliography{ref}